\documentclass[10pt,letterpaper]{article}
\usepackage[top=0.85in,left=2.75in,footskip=0.75in]{geometry}

\usepackage{changepage}

\usepackage[utf8]{inputenc}

\usepackage{textcomp,marvosym}

\usepackage{fixltx2e}

\usepackage{amsmath,amssymb}

\usepackage{cite}

\usepackage{nameref,hyperref}

\usepackage{microtype}
\DisableLigatures[f]{encoding = *, family = * }

\usepackage{rotating}
\usepackage{color}

\usepackage{hyperref}

\raggedright
\usepackage[aboveskip=1pt,labelfont=bf,labelsep=period,justification=raggedright,singlelinecheck=off]{caption}

\makeatletter
\renewcommand{\@biblabel}[1]{\quad#1.}
\makeatother

\date{}

\usepackage{lastpage,fancyhdr,graphicx}
\usepackage{epstopdf}
\fancyheadoffset[L]{2.25in}
\fancyfootoffset[L]{2.25in}
\newcommand {\ea} {{\it et al.}}
\newcommand{\argmax}{\operatornamewithlimits{argmax}}
\begin{document}
\vspace*{0.35in}

% Title must be 150 characters or less
\begin{flushleft}
{\Large
\textbf\newline{A Deep-structured Conditional Random Field Model for Object Silhouette Tracking}
}
% Insert Author names, affiliations and corresponding author email.
\newline
M.~J. Shafiee$^{1,\ast}$,
Z. Azimifar$^{2}$,
A. Wong$^{1}$
\\
\bf{1} Department of Systems Design Engineering, University of Waterloo, Waterloo, Ontario, Canada
\\
\bf{2} Department of Computer Science and Engineering, Shiraz University, Shiraz, Fars, Iran
\\
\bigskip
M.S., Z.A., and A.W. conceived and designed the DS-CRF model.  M.S., Z.A., and A.W. worked on formulation and derivation of solution for the DS-CRF model.  M.S. performed the data processing.  All authors contributed to writing the paper and to the editing of the paper.

$\ast$ E-mail: mjshafiee@uwaterloo.ca
\\

\end{flushleft}

% Please keep the abstract between 250 and 300 words
\section*{Abstract}
In this work, we introduce a deep-structured conditional random field (DS-CRF) model for the purpose of state-based object silhouette tracking.  The proposed DS-CRF model consists of a series of state layers, where each state layer spatially characterizes the object silhouette at a particular point in time.  The interactions between adjacent state layers are established by inter-layer connectivity dynamically determined based on inter-frame optical flow.  By incorporate both spatial and temporal context in a dynamic fashion within such a deep-structured probabilistic graphical model, the proposed DS-CRF model allows us to develop a framework that can accurately and efficiently track object silhouettes that can change greatly over time, as well as under different situations such as occlusion and multiple targets within the scene. Experiment results using video surveillance datasets containing different scenarios such as occlusion and multiple targets showed that the proposed DS-CRF approach provides strong object silhouette tracking performance when compared to baseline methods such as mean-shift tracking, as well as state-of-the-art methods such as context tracking and boosted particle filtering.

%\linenumbers

\section*{Introduction}

Structured prediction, where one wishes to predict structured states given structured observations, is an interesting and challenge problem that is important in a number of different applications, with one of them being object silhouette tracking.  The goal of object silhouette tracking is to identify the silhouette of the same object over a video sequence, and is very challenging due to a number of factors such as occlusion, object motion changing dynamically over a video sequence, and object silhouette changing drastically over time.

Much of early literature in object tracking have consisted of generative tracking methods, where the joint distribution of states and observations is modeled.  The classical example is the use of Kalman filters~\cite{Kalman}, where predictions of the object are made with Gaussian assumptions made on both states and observations based on predefined linear system dynamics.  However, since object motion do not follow Gaussian behaviour and have non-linear system dynamics, the use of Kalman filters can often lead to poor prediction performance for object tracking.  To address the issue of non-linear system dynamics, researchers have made use of modified Kalman filters such as the extended Kalman filter~\cite{EKF1,EKF2} and unscented Kalman filter~\cite{UKF}, but these do not resolve issues associated with non-Gaussian behaviour of objection motion.  To address both the issue of non-linear system dynamics and non-Gaussian behaviour, a lot of attention has been paid to the use of particle filters~\cite{particlefilter1,particlefilter}, which are non-parametric posterior density estimation methods that can model arbitrary statistical distributions.  However, the use of particle filters for object tracking is not only computationally expensive, but difficult to learn especially for the case of object silhouette tracking where motion and silhouette appearance can change drastically and dynamically over time.

Recently, there has been significant interest in the use of discriminative methods for object tracking over the use of generative methods.  In contrast to generative methods, discriminative methods directly model the conditional probability distribution of states given observations, and relax the conditional independence assumption made by generative methods.  In particular, conditional random fields (CRF) are the most well-known discriminative graphical models used for the purpose of structured prediction, and have shown in a large number of studies to outperform generative models such as \mbox{Markov random fields \cite{CRFMRF}}.  Motivated by this, a number of CRF-based methods have been proposed for the purpose of object tracking.  Taycher~\ea~\cite{related1} proposed a human tracking approach using CRFs, with an $L_1$ similarity space corresponding to the potential functions. Different poses were considered as tracked states within a video sequence, where as the number of states must be predefined by the user. Sigal~\ea~\cite{related4} used two-layer spatio-temporal models for component-based detection and object tracking in video sequences. In that work, each object or component of an object was considered as a node in the graphical model at any given time. Moreover, the graph edges correspond to learned spatial and temporal constraints. Following this work, \mbox{Ablavsky~\ea~\cite{related5}} proposed a layered graphical model for the purpose of partially-occluded object tracking.  A layered image plane was used to represent motion surrounding a known object that is associated with a pre-computed graphical model.  CRFs have also been applied to image-sequence segmentation~\cite{related3,related2}, where the random fields are modeled using spatial and temporal dependencies.  \mbox{Shafiee \ea~\cite{javad}} proposed the concept of temporal conditional random fields (TCRF) for the purpose of object tracking, where the object's next position is estimated based on the current video frame, and then subsequently refined via template matching based on a subsequent video frame.

The use of CRFs specifically related to object silhouette tracking is more recent and as such more limited in existing literature.  Ren and Malik~\cite{Ren} proposed the use of CRFs for object silhouette segmentation in video sequences where the background and foreground distributions are updated over time.  In the work by Boudoukh~\ea\cite{CRFVisualSilho}, a target silhouette is tracked on a video sequence by fusing different visual cues through the use of a CRF.  In particular, temporal color similarity, spatial color continuity, and spatial motion continuity were considered as the CRF feature functions.  The key advantage of this method for object silhouette tracking is that pixel-wise resolution can be achieved.

While such CRF-based approaches to object silhouette tracking shows significant promise, one inherent limitation that is faced is that the existing CRF models used to predict the object silhouettes for one video frame are limited in their ability to take greater advantage of information from other video frames.  One can use more complex CRF models to increase modeling power to address these limitations for improved object silhouette tracking, but it would also significantly increase computational complexity as well as model learning complexity.  Recently, the concept of deep-structured  models have been proposed to facilitate for increased modeling power without the significant increase in computational complexity and model learning complexity incurred by complex CRF models.  Deep-structured CRF models make use of intermediate state layers to improve structured prediction performance, where there is an inter-layer dependency between each layer on its previous layer.  Ratajczak et al.~\cite{Ratajczak} proposed a context-specific deep CRF model where the local factors in linear-chain CRFs are replaced with sum-product networks.  Yu et al.~\cite{Yu1,Yu2} proposed a deep-structured CRF model composed of multiple layers of simple CRFs, with each layer's input consisting of the previous layer's input and the resulting marginal probabilities.  Given that the problem of object silhouette tracking is one where a set of video frames can contribute to predicting the object silhouette in a new video frame, one is motivated to investigate the efficacy of deep-structured CRF models for solving this problem.

In this work, we propose an alternative framework for state-based object silhouette tracking based on the concept of deep-structured discriminative modeling.  In particular, we introduce a deep-structured conditional random field (DS-CRF) model consisting of a series of state layers, with each state layer spatially characterizes the object silhouette at a particular point in time.  The interactions between adjacent state layers are established by inter-layer connectivity dynamically determined based on inter-frame optical flow.  By incorporate both spatial and temporal context in a dynamic fashion within such a deep-structured probabilistic graphical model, the proposed DS-CRF model allows us to develop a framework that can accurately and efficiently track object silhouettes that can change greatly over time.  Furthermore, such a modeling framework does not require distinct stages for prediction and update, and does not require independent training for the dynamics of each object silhouette being tracked.  Experimental results show that the proposed framework can estimate object silhouettes over time in situations where there is occlusion as well as large changes in object silhouette appearance over time.

\section*{Materials and Methods}
Within a statistical modeling framework, one can describe the problem of object silhouette tracking as a classification problem, where the goal is to classify each pixel in a video frame as either foreground (part of the object silhouette) or background. The goal is to maximize the posterior probability of the states given observations $P(Y|M)$, where, $Y$ is the state plane characterizing the object silhouette and $M$ is corresponding observations (e.g., video).  Discriminative models, such as CRFs, derive the posterior probability $P(Y|M)$ directly and as such do not require the independence assumptions necessary for generative modeling approaches.  In the proposed DS-CRF modeling framework for object silhouette tracking, the object silhouette and corresponding background at the pixel level for each video frame is characterized by a state layer, which the series of state layers interconnected based on inter-frame optical flow information to form a deep-structured conditional random field model that facilitates for interactions amongst adjacent state layers.  A detailed description of CRFs in the context of object silhouette tracking, followed by a detailed description of the proposed DS-CRF model, is provided below.

\subsection*{Conditional Random Fields}
\label{CRF}
Conditional random fields (CRFs) are amongst the most effective and widely-used discriminative modeling tools developed in the past two decades. The idea of CRF modeling was first proposed by \mbox{Laffety~\ea~\cite{CRF}}; based on the Markov property, the CRF directly models the conditional probability of the states given the measurements, without requiring the specification of any sort of underlying prior model, and relaxes the conditional independence assumption commonly used by generative models.

Formally, let $G = (V,E)$ be an undirected graph such that $y_i\in Y$ is indexed by the vertices of $v_i \in V$ in $G$. $(Y,M)$ is said to be a CRF if, when globally conditioned on $M$, the random variables $y_i$ obey the Markov property with respect to the graph $G$. In other words, $P(y_i|M,y_{V-\{i\}}) = P(y_i|M,y_{N_i})$ where ${V-\{i\}}$ is the set of all nodes in $G$ except node $i$, $Y$ is a set of output variables that we aim to predict, and $N_i$ and $M$ are the sets of neighbors of $i$ and of observed input variables, respectively. The general form of a CRF is given by
\begin{eqnarray}
P(Y|M)= &\frac{1}{Z(M)}
\prod _{c\in C} \psi_c(y_c,M_c)
\\ Z(M) = &\sum_y \prod _{c\in C} \psi_c(y_c,M_c) \nonumber
\label{CRFgeneraleq}
\end{eqnarray}
\noindent where $Z(M)$ is a normalization constant, essentially the so-called partition function of Gibbs fields, with respect to all possible values of $Y$, $C$ represents the set of all cliques, and $\psi_c$ encodes potential functions with a non-negative value condition.

According to the non-negative constraint for $\psi_c$, and based on the Principle of Maximum Entropy~\cite{Jaynes}, a proper probability distribution is the one that maximizes the entropy, given the constraints from the training set \cite{dortmond}.  As such, a new form of the CRF is then given by
\begin{align}
\label{CRFeq}
P(Y|M)=  \frac{1}{Z(M)}
\prod _{c\in C} \prod _{\phi_c\in c} \exp\Big( \sum_k \lambda _{\phi_c,k} f_{{\phi_c,k}}  (Y_{\phi_c},M) \Big)
\end{align}
\noindent where $ f_{{\phi_c}}(Y_{\phi_c},M)$ is a feature function with respect to clique $\phi_c$, and $\lambda$ denotes the weight of each feature function to be learned.  The feature function expresses the relationship amongst the random variables in a clique.  The number of feature functions with respect to each clique is denoted by $k$.

Two-dimensional CRFs have been applied to many computer vision problems, such as segmentation and classification. In particular, because of the undirected structure of most images, the 2D CRF leads to efficient performance in computer vision~\cite{CRFMRF,sCRF1,scrf2}.  Although early CRFs incorporate spatial relationships (spatial feature functions) amongst random variables into the model, these relationships repeat sequentially in many applications such as visual tracking, where incorporating this property into the framework can lead to better modeling.

\begin{figure}[!h]
\begin{center}
  \includegraphics[scale = 0.4]{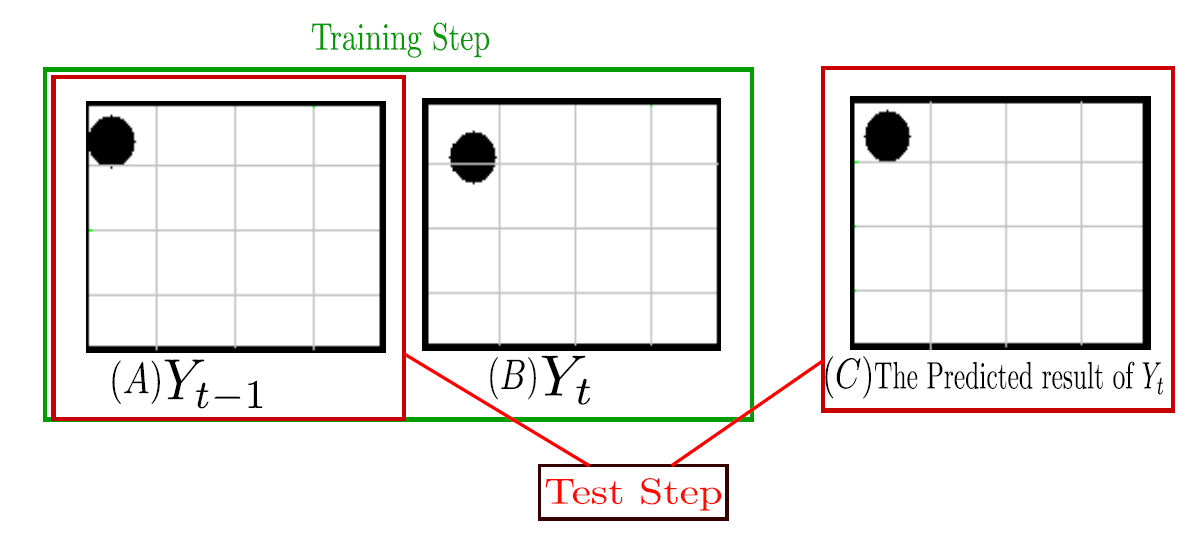}
\end{center}
\vspace{-0.3 cm}
   \caption{\textbf{Example of CRF modeling of object motion using only spatial feature functions for object silhouette tracking.} The first column (A) shows the temporal observation and second column (B) shows the label used for training (two columns are two consecutive frames at $t-1$ and $t$). The third column (C) shows the prediction result of the object silhouette.  As seen  the prediction result of the
object silhouette is poor since the CRF could not learn object motion dynamics in the absence of temporal feature functions, leading to poor object silhouette tracking performance.}
\label{fig:CRFincon}
\end{figure}

Feature functions play an important role in the context of CRF modeling. Selecting appropriate feature functions speeds up the convergence of the CRF training process, whereas inappropriate feature functions can cause inconsistent results in CRF inference.  To illustrate the importance of selecting appropriate feature functions for object silhouette tracking, we train a CRF for predicting the object silhouette at one frame based on the  previous frame using only spatial feature functions without incorporating any feature function describing temporal relationship amongst frames. Two frames consist of a simulated object which has  small movement between two frames.  As seen in Fig.~\ref{fig:CRFincon}, the prediction result of the object silhouette is poor as the CRF could not learn object motion dynamics in the absence of temporal feature functions, leading to poor object silhouette tracking performance.

To tackle this issue of selecting appropriate feature functions to improve tracking performance, Shafiee~\ea~\cite{javad} proposed the incorporation of temporal feature functions such as inter-frame optical flow into the CRF modeling framework to better take advantage of temporal relationships for visual tracking. Although this approach showed promising results and illustrated the feasibility of temporal processing for visual tracking in the CRF modeling framework, it only makes use of motion information from the previous frame to estimate object position in the current frame and as such cannot handle large motion dynamics changes nor shape changes over time, or can it handle accelerated motion dynamics.  Furthermore, it is designed for object position tracking and does not handle object silhouette tracking.  Therefore, motivated by the benefits of incorporating both spatial and temporal context in a dynamic fashion in a manner that addresses the aforementioned issues, we propose a deep-structured CRF (DS-CRF) model for object silhouette tracking, where the series of interconnected state layers making up the model along with the set of corresponding temporal observations allow for better modeling of more complex motion and shape dynamics that can occur in realistic scenarios.

\subsection*{Deep-structured Conditional Random Fields}
\label{Sec-CRFP}
%\vspace{-0.3 cm}
Here, we will describe the proposed DS-CRF model in detail as follows.  First, the graph representation of the DS-CRF model is presented.  Second, the manner in which inter-layer connectivity within the DS-CRF model is established dynamically based on motion information derived from temporal observations is presented.  Third, a set of new feature functions incorporated in the DS-CRF model for object silhouette tracking is presented.

\begin{figure*}[!h]
\begin{center}

   \includegraphics[scale = 0.35]{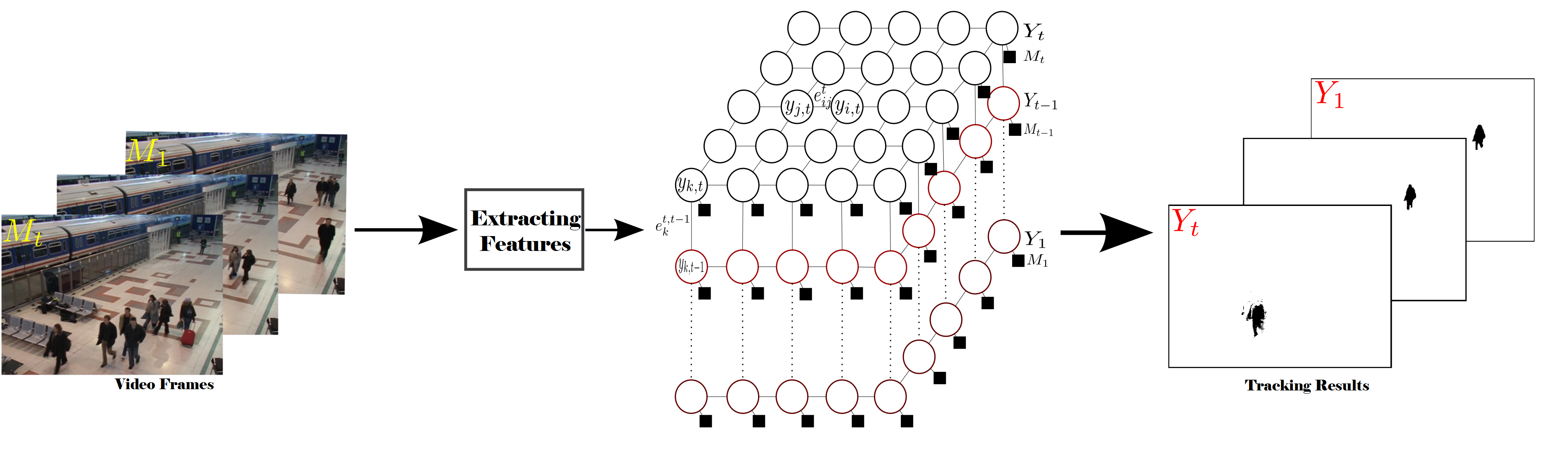}
\end{center}
%\vspace{-0.3 cm}
\caption{\textbf{The flow diagram of the proposed framework.}  The observed frames are utilized to predict the new target position. Several feature functions  are incorporated  into the model.  The extracted features  from the observed frames are utilized as the measurement layer while the new object location inferred in a black and white plane is considered as the label field (tracking result) in the random field. }
\label{fig:flowchart}
\end{figure*}

Fig.~\ref{fig:flowchart} demonstrates the flow digram of the proposed framework.  Several features such as optical flow are extracted from observed frames to track the new target location in the video. The tracking result is encoded as a black and white field demonstrating the target location with black pixels.

\subsection*{Graph Representation}

\begin{figure*}[!h]
\begin{center}

   \includegraphics[scale = 0.75]{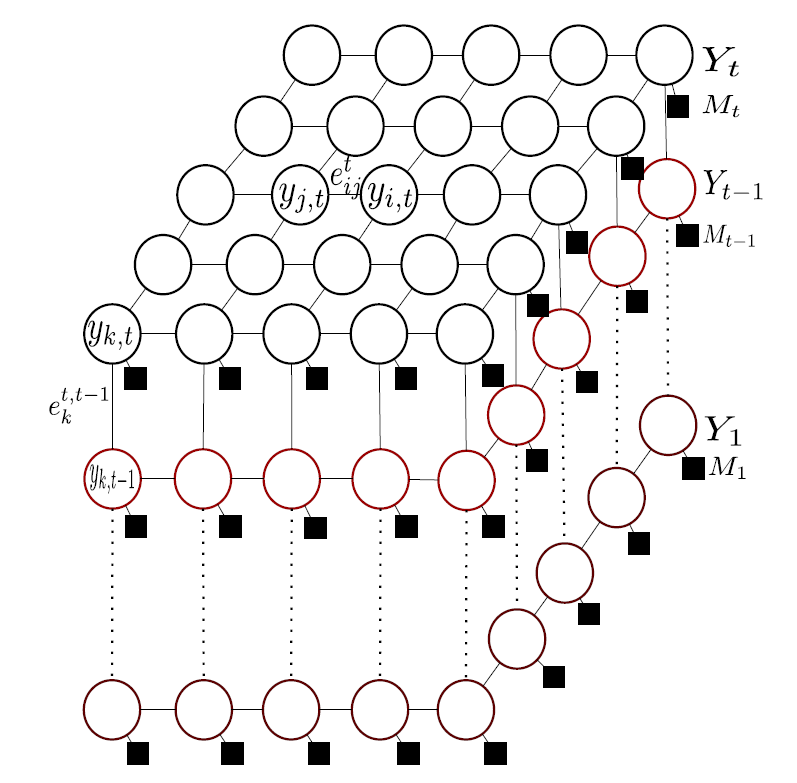}
\end{center}
%\vspace{-0.3 cm}
\caption{\textbf{Graph representation of the deep-structured conditional random field model.}  In this model, the labels $Y_{t-1} : Y_1$ and all observations can be incorporated to model the object silhouette's motion.  Two different types of clique connectivity exist in this model: i) inter-layer clique connectivity between two nodes within the same layer as shown by $e_{ij}^t$, and ii) inter-layer clique connectivity between nodes in two adjacent layers.  The two end-nodes of an inter-layer clique are determined based on the motion information.}
\label{fig:DeepModel}
\end{figure*}

Let the graph $G(V,E)$ represent the proposed DS-CRF model, which consists of several state layers $Y_t:Y_1$ corresponding to times $t:1$ as shown in Fig.~\ref{fig:DeepModel}.  Each state layer characterizes the object silhouette at a specific time step by modeling the conditional probability of $Y_t$ given the previous states of the object in times $t-1$:1 and their corresponding observations:
\begin{align}
P(Y_{t}|M_{t:1},Y_{t-1:1})=
\frac{1}{Z(M_{t:1})}  \prod _{c \in C} \exp \Big (\sum_k \lambda _{k,c} f_{{k,c}}  (Y_{c,t},M_{t:1},Y_{t-1:1}) \Big ).
\label{CRFP1}
\end{align}
\noindent where $Z(M_{t:1})$ is a normalization constant, $C$ is the set of inter-layer and intra-layer cliques, $\lambda _{k,c}$ determines the weight of each feature function, and $f_{{k,c}}  (\cdot)$ denotes the feature function over clique $c$.  The intra-layer connectivity between nodes in each layer (i.e, $e^t_{ij}$ in layer $Y_t$, Fig.~\ref{fig:DeepModel}) imposes the smoothness property of the target object into the model while the inter-layer connectivity between two adjacent state layers (i.e., $e^{t,t-1}_k$ for layers $Y_t$ and $Y_{t-1}$ corresponding to node $y_k$)  incorporate object motion dynamics into the model.  As such, the inter-layer connectivity carries the energy corresponding to unary potential in the model, and are specified dynamically and adaptively in the proposed framework based on motion information derived from temporal observations, which will be described in detail in the next section.

To reduce computational complexity, the implementation of the DS-CRF model in this work will make use of the three previous frames as observations in the modeling of the conditional probability:
\begin{align}
P(Y_{t}|M_{t-2:t},Y_{t-1:t-2})=
\frac{1}{Z(M_{t:t-2})}  \prod _{c \in C} \exp \Big ( \sum_k \lambda _{k,c} f_{{k,c}}  (Y_{c,t},M_{t:t-2},Y_{t-1:t-2}) \Big ).
\label{CRFP}
\end{align}
\noindent The use of the last three frames is chosen as it provides sufficient information to reasonably model accelerated motions given that both acceleration and velocity can be computed, thus allowing for handling various motion situations in short time steps.

\subsection*{Motion-guided inter-layer connectivity}
As discussed above, the intra-layer connectivity between nodes in a layer incorporate spatial context while the inter-layer connectivity between layers incorporate temporal context into the DS-CRF model.  The simplest approach to establishing inter-layer connectivity between nodes from different layers would be to simply create inter-layer cliques between nodes that represent the same spatial location at two different time steps.  This creates a simple regular spatial-temporal lattice that is fixed across time.  However, this is not appropriate for object silhouette tracking, as temporal neighbors established under such a fixed inter-layer connectivity structure would not share relevant information since target objects that are undergoing drastic motion and shape changes over time, and thus the feature functions under such a structure would hold little meaning.  Therefore, we are motivated to establish the inter-layer connectivity in the propose DS-CRF model in a dynamic and adaptive manner, where motion information derived from the temporal observations is used to determine the inter-layer cliques at each state layer.

In this work, we dynamically determine inter-layer cliques of each node at each layer $Y_t$ of the DS-CRF model based on the velocity obtained by inter-frame optical flow computed by two consecutive temporal observations $M_t$ and $M_{t-1}$:
\begin{align}
y_{c,t} = \Big\{(y_{i,t}, y_{k,t-1})| k_x-v_x = i_x,k_y-v_y = i_y \Big\}
\end{align}
where $y_{c,t}$ is an inter-layer clique in time $t$, $y_{i,t}$ is a node in time $t$ and $y_{k,t-1}$ is its neighbor node in time $t-1$ based on the inter-layer clique connectivity. $v_x$ and $v_y$ encode the velocities in both directions of $x$ and $y$ where node $i$ in $x$ direction (i.e., $i_x$) is consistent with node $k$ (i.e., $k_x$) based on $v_x$ and the same manner for $y$ direction.

\begin{figure*}[!h]
\begin{center}

  \includegraphics[scale = 0.85]{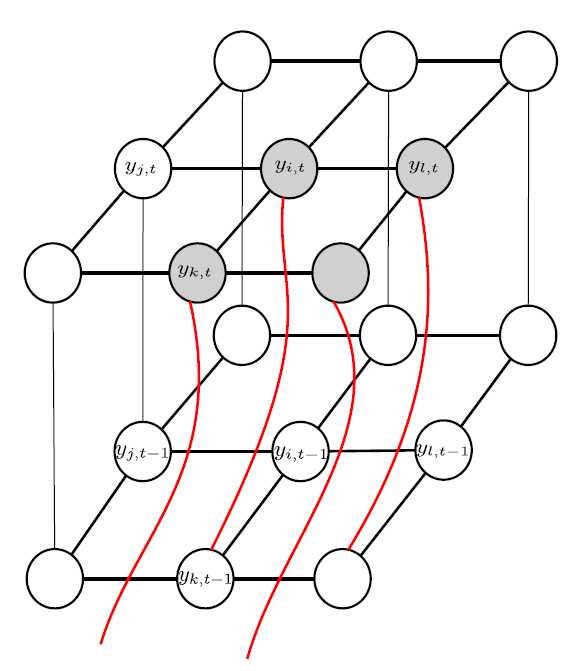}
\end{center}
\vspace{-0.3 cm}
\caption{\textbf{Inter-layer connectivity.} The inter-layer connectivity between nodes in two adjacent layers are determined based on the motion information computed in two consecutive frames. The inter-layer cliques are constructed dynamically and adaptively based on motion information corresponding to each node in layer $Y_t$. As seen, the corresponding temporal neighbor of node $y_{i,t}$ is $y_{k,t-1}$ based on inter-layer clique structure are determined by use of inter-frame optical flow.  The gray color correspond to nodes associated with the target object where there is movement in the previous frame.}
\label{fig:displacementclique}
\end{figure*}

An illustrative example of the motion-guided inter-layer connectivity strategy is shown in Fig.~\ref{fig:displacementclique}, where the inter-layer clique structures are established at $Y_t$ and $Y_{t-1}$ based on the inter-frame optical flow between temporal observations $M_t$ and $M_{t-1}$.  It can be seen that the nodes corresponding to the target object (indicated here as gray nodes) (e.g., $y_{i,t}$, $y_{l,t}$, and $y_{k,t}$) form inter-layer clique structures with nodes from the previous state layer that characterize different spatial locations than them due to motion, while nodes corresponding to the background (indicated by white nodes) (e.g., $y_{j,t}$) form inter-layer cliques with nodes from the previous state layer corresponding to the same spatial location since there is no motion at that position.  As such, this motion-guided dynamic inter-layer connectivity strategy allows for better characterization of temporal context of the object silhouette being tracked and allow the feature functions to hold meaning.

\subsection*{Feature Functions}
In addition to the inter-layer connectivity between state layers, it is important to also describe the feature functions being incorporated into the proposed DS-CRF model.  The set of feature functions are: i) optical flow, ii) target appearance, iii) spatial coherency, and iv) edge.\\

\noindent \textbf{Optical Flow}
This crucial feature function is described by the velocity of each pixel in the $x$ and $y$ directions in two adjacent frames and is estimated via inter-frame optical flow.  Optical flow is an approximation of motion based upon local derivatives in a given sequence of images \cite{optflow}. It specifies the moving distance of each pixel in two adjacent images:
\begin{align}
I(x,y,t) = I((x+\delta x),&(y+\delta y),(t+\delta t)) \nonumber \\
(I_x,I_y)\cdot(v_x,v_y) &= -I_t
\end{align}
\noindent where $(I_x,I_y)$ denotes the spatial intensity gradient and $v_x$ and $v_y$ denotes motion in both directions (here, in this implementation, $\delta t = 1$).  Optical flow assumes the change in a pixel's intensity corresponds to the displacement of pixels~\cite{javad}.  Here, inter-frame optical flow is applied between two temporal observations.\\~\\
\noindent \textbf{Target Appearance}
The model utilizes simple unary appearance feature functions based on features describing the target object's appearance, including RGB color and  target appearance in previous frame. To obtain this unary feature function, the label state of time $t-1$ is shifted by the computed velocity and find the corresponding value for each node:
\begin{align}
f(y_{i,t},M) =  \mathcal{S}(Y_{t-1},v_x,v_y)
\end{align}
where $\mathcal{S}(\cdot)$ shifts $Y_{t-1}$ based on velocities $v_x$ and $v_y$.\\~\\
\noindent \textbf{Spatial Coherency}
Each target in the scene has spatial color coherency. This term implies the reflection between neighboring nodes in the image. Each node consisted to a target has strong relations with other nodes corresponding to the target silhouette. In other words, the target appearance is coherent in each time frame. By adding this feature function to the DS-CRF tracking framework, the proposed algorithm can track target object's silhouette despite large changes over time. A rough segmentation algorithm~\cite{nock2004statistical} enforces the label consistency among nodes with a segment produced by the segmentation result of frame $M_t$.\\~\\
\noindent \textbf{Edge}
 The Ising model is the ordinary edge energy function utilized in different problems which we incorporated to the model as spatial smoothness feature function:
 \begin{align}
 f(y_{i,t},y_{j,t}, M_t) = \mu(y_{i,t},y_{j,t}) \cdot (m_{i,t} - m_{j,t})
\end{align}
\noindent where $\mu(\cdot)$ is the penalty function based on the similarity of two nodes $i$ and $j$.

\subsection*{Training and Inference}
\label{tran_infer}
Maximum likelihood is a common method to estimate the parameters of CRFs. As such, the training of the proposed DS-CRF is done by maximizing log-likelihood ${\ell}$ upon the training data:
\begin{eqnarray}
\label{logLikelihood}
    \ell(\lambda)= \sum _{c \in C} \Big (\sum_k \lambda _{k,c} f_{{k,c}}  (Y_{c,t},M_{t-2:t},Y_{t-1}) \Big ) - log\Big(Z(M_{t-2:t})\Big).
\end{eqnarray}
\noindent Because the log-likelihood function $\ell(\lambda)$ is concave, the parameters $\lambda$ can be chosen such that the global maximum is obtained and the gradient or vector of partial derivatives with respect to each parameter $\lambda_{k}$ becomes zero. Differentiating $\ell(\lambda)$ with respect to the parameter $\lambda_{k}$ gives:

\begin{eqnarray}
 \frac{\partial\ell}{\partial\lambda_{k}}=\sum_{c \in C}\Big(f_{{k,c}}  (Y_{c,t},M_{t-2:t},Y_{t-1})
 -\sum_{Y'} f_{{k,c}}  (Y_{c,t},M_{t-2:t},Y_{t-1}) P(Y'|X )\Big).
\end{eqnarray}
An exact solution does not exist; therefore, the parameters are determined iteratively using gradient descent optimization. Our DS-CRF training is performed via the belief propagation method \cite{CRFtraining}.

After the training of the DS-CRF, inference is performed by evaluating the probability of each random variables in the represented graph given the observations $M_{t-2:t-1}$ and $Y_{t-1}$, while decoding is performed by assigning the output variable $Y$ --- determining states with maximum probability:
\begin{equation}\label{infer}
P_{y_i} = P(Y_{t} = y_i |M_{t-2:t-1}, Y_{t-1})  \text{           }  \forall  y_i \in Y
\end{equation}
\begin{equation}\label{decoding}
Y^{*}=\argmax_{Y}P(Y_{t}|M_{t-2:t-1}, Y_{t-1}).
 \end{equation}
where Eq.~\eqref{infer} and Eq.~\eqref{decoding} show the formal definition of the inference and decoding process, respectively.

\subsection*{DS-CRF Tracking Framework}
\label{CRF_TFramework}
Based on the DS-CRF model described above, we can then develop a state-based framework for tracking object silhouettes across time in a video sequence as follows.
The first two frames are annotated by user as initialization. The velocity is computed based on these two frames and DS-CRF starts the tracking procedure by third frame.  DS-CRF can track objects automatically after frame 2.   The optical flow is performed by used on two last seen frames each time. Since the optical flow is computed for each time frame and parameters were trained based on the velocity, the model needs to train only one time.

   The DS-CRF essentially plays the rule of fusing spatial context such as target object shape and appearance with temporal context such as motion dynamics within the proposed tracking framework.  The contribution of each aspect of the spatial and temporal information within the DS-CRF model based on the weights learned during the training step.  The inference process described in Eq.~\ref{infer} is then performed based on the learned DS-CRF.

\begin{figure}[!ht]
\begin{center}

   \includegraphics[scale = 0.8]{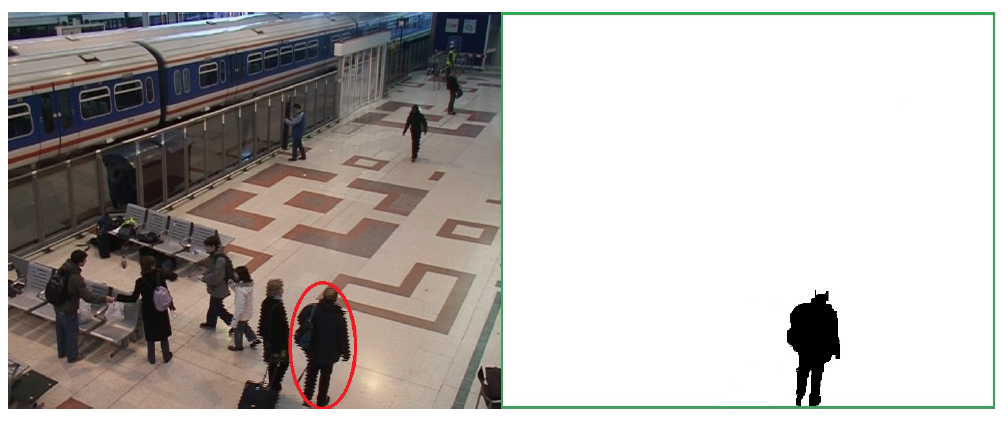}
\end{center}

\caption{\textbf{ An example of a temporal observation and the corresponding label field.} The red oval indicates the target object being tracked.}
\label{fig:ml}
\end{figure}

To reduce the computational complexity of the proposed DS-CRF tracking framework, the decoding result in each step~(see Eq.~\ref{decoding}) utilized to obtain the object silhouette.  The decoding result consists of a binary label field $Y$ (i.e., each pixel has a value $y = \{0,1\}$, where $0$ indicates object silhouette pixels and $1$ indicates background pixels).  An example of a temporal observation (i.e., video frame) and its corresponding binary label field is shown in Fig.~\ref{fig:ml}.  The use of a binary label setup allows for not only reduced computational complexity of the training process, but also the convergence.

One issue that needs to be tackled when using a binary label setup is that, while well suited for single target object silhouette tracking, it is less appropriate for multi-target object silhouette tracking.  To address this issue, we introduce a data association procedure where connected components in the binary label field are assigned to the target objects being tracked.  This is accomplished by matching the object silhouettes determined for the previous time step to the connected components in the binary field at the current time step to determine the best template matches:
\begin{align}
T_j(t) = \underset{c_i \in \mathcal{C}}{argmax}  \mathcal{M}(c_i,T_j(t-1))
\end{align}
where $T_j(t)$ is the target $j$'s silhouette in time $t$, $\mathcal{C}$ is the set of connected components  detected as targets and $\mathcal{M}(\cdot)$ is template matching function that evaluates the similarity of two input silhouettes.

\section*{Results}

To evaluate the performance of the proposed DS-CRF model for the purpose of object silhouette tracking, a number of different experiments were performed to allow for a better understanding and analysis of the model under different conditions and factors.   First, a set of experiments involving video of a simulated object with different motion dynamics is performed to study the capability of the DS-CRF model in handling objects with changing motion dynamics.  Second, a set of experiments performed on videos of humans moving within a subway station from the PETS2006 database is used to study the capability of the DS-CRF model in handling object silhouette tracking scenarios where there is occlusion and objects that change drastically in shape and size over time.

\subsection*{Experiment 1: Simulated object motion undergoing acceleration}
\label{CRFP-R}
In this experiment, we examine the ability of the proposed DS-CRF method in tracking the silhouette of an object with different motion dynamics over time.  To accomplish this, we produce three video sequences consisting of a simulated object undergoing the following motion dynamics over time:
\begin{itemize}
\item Motion1: Object undergoes acceleration but remain constant in shape over time.
\item Motion2: Object undergoes size change over time but moves at constant velocity.
\item Motion3: Object undergoes acceleration as well as size change over time.
\end{itemize}

\begin{figure}[!h]

\begin{center}

 \includegraphics[scale = 1.7]{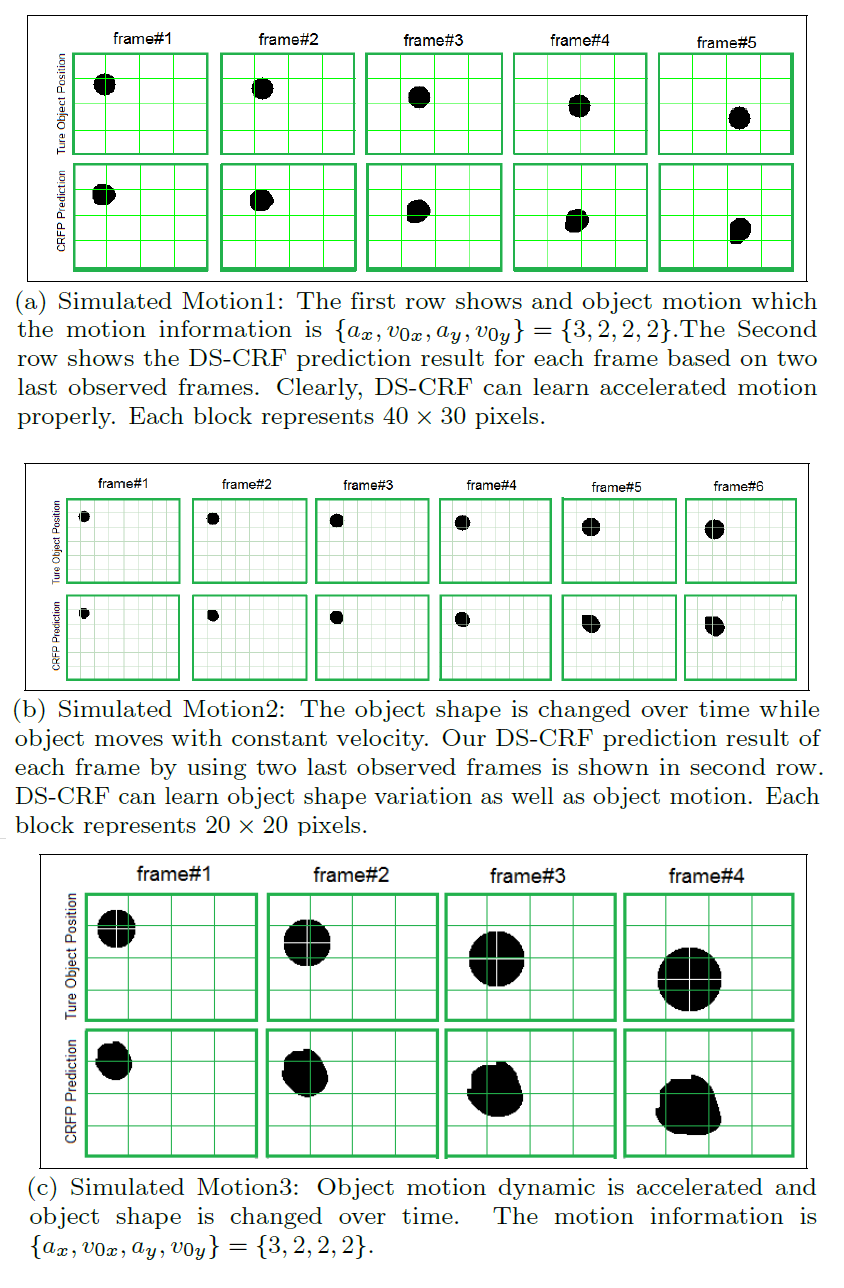}
\end{center}
\caption{\textbf{Examined simulated motion results.} DS-CRF is examined with three different simulated motion and shape dynamics over time.  In (a) the object undergoes acceleration but remain constant in shape over time, in (b) the object undergoes size change over time but moves at constant velocity, and in (c) the object undergoes acceleration as well as size change over time. }
\label{fig:non-rigid_all}
\end{figure}

  Sample frames from the video sequences Motion1, Motion2, and Motion3 is shown in the first rows of \mbox{Fig.~\ref{fig:non-rigid_all}}(a),(b), and (c), respectively.  The proposed method was then used to predict the object silhouette over time based on these video sequences.  The predicted results are shown in the second rows of \mbox{Fig.~\ref{fig:non-rigid_all}}(a),(b), and (c), respectively.  It can be observed that the proposed method is able to provide accurate object silhouette tracking results for all three video sequences, thus illustrate its ability to handle uncertainties in both motion and object appearance over time.

\subsection*{Experiment 2: Real-life video of human targets}
\label{CRFT-R}

In this experiment, we examine the capability of the DS-CRF model in handling object silhouette tracking scenarios where there is occlusion and objects that change drastically in shape and size over time.  To accomplish this, we made use of three different video sequences from the PETS2006 database depicting human targets moving within a subway station (one of which is used for evaluation in~\cite{CRFVisualSilho}), each used to illustrate different aspects of the capability of the proposed method:

\begin{itemize}
\item Subway1: This sequence is used to illustrate the capability of the proposed method in handling single object silhouette tracking over time.  The object target in this sequence is crossing the hallway from the top of the scene to the bottom of the scene.
\item Subway2: This sequence is used to illustrate the capability of the proposed method in handling object occlusions. The object target in this sequence is crossing the hallway from the right of the scene to the left of the scene, and becomes occluded by a person walking from the left of the scene to the right of the scene.
\item Subway3: This sequence is used to illustrate the capability of the proposed method in handling multiple object silhouette tracking over time.  Two of the target objects  in this sequence is crossing the hallway from the bottom of the scene to the top of the scene, while a third object target is crossing the hallway from the top of the scene to the bottom of the scene.
\end{itemize}
The PETS2006 database is a public dataset which is available from \url{http://www.cvg.reading.ac.uk/PETS2006/data.html}

To provide a comparison for the performance of the proposed method, four different existing tracking methods are also evaluated:

\begin{itemize}
\item {Mean-shift tracking~\cite{meanshift}}
\label{MS}
Mean-shift tracking is based on non-parametric feature space analysis, where the goal is to determine the maxima of a density function, which in the case of visual tracking is based on the color histogram of target object.  This goal is achieved via an iterative optimization strategy that locates the new target object position near the previous object position based on a similarity measure such as Bhattacharyya distance.

\item {Context tracking~\cite{context}}
\label{CVPR2011}
 Context tracking is a discriminative tracking approach which utilizes a specific trained detector in a semi-supervised fashion to locate the target in consecutive frames.  The goal of this method is to locate all possible regions that look similar to the target.  Context tracking then identifies and differentiates the target object from the `distracters' within the set of possible regions based on a confidence measure derived based on the posterior probability and supporting features.

\item {Boosted particle filtering~\cite{particlefilter}}
\label{PF}
Particle filtering is a discriminative tracking approach that approximates the posterior $P(Y_t|M_{0:t})$ with a Dirac measure using a finite set of $N$ particles $\{Y^i_t\}_{i = 1...N}$. The sample candidate particles are drawn based on the proposal distribution. The importance weight of each particle is then updated according to  its previous weight and the importance function, which is often the transition prior.  After that, the particles are re-sampled using their importance weights.  Here, we employed the boosted particle filter proposed in~\cite{particlefilter}, which incorporates mixture particle filtering~\cite{mixedpf} that is ideally suited to multi-target tracking.

\item {Visual Silhouette Tracker \cite{CRFVisualSilho}}
\label{VST}
The visual silhouette tracking method fuses different visual cues by means of conditional random fields.  The object silhouette is estimated every frame according to visual cues including temporal color similarity, spatial color continuity and spatial motion continuity. The incorporated energy functions are minimized within a conditional random field framework.
\end{itemize}

Note that for the mean-shift tracking and context tracking methods are only evaluated for the Subway1 and Subway2 sequences as the implementations used were not designed for tracking multiple object targets within the same scene.  The visual silhouette tracking method was only compared for the Subway1 sequence as only the object silhouette results for that sequence was provided by the authors of~\cite{CRFVisualSilho}.  Finally, the boosted particle filtering and proposed DS-CRF method was evaluated for all three sequences (Subway1, Subway2, and Subway3).

To compare  methods quantitatively,  the number of frames which the tracker could track  the object correctly divided by the total number of frames in the sequence is reported as the accuracy:
\begin{align}
Accuracy = \frac{\text{Number of Corrected Tracked Frames}}{\text{Total Number of Frames}}\times 100.
  \end{align}
      Table~\ref{tab:res1} shows the quantitative results for  the Subway1 and Subway2 sequences  while Table~\ref{tab:res2} presents the result corresponding to Subway3 sequence.

\begin{table*}[!Htp]
	
	\begin{center}
    \caption{Quantitative results for different video sequences. The accuracy of Visual Silhouette Tracker method is  reported   for one sequence since only one sequence of result has been provided by the authors of~\cite{CRFVisualSilho}. MST, CT, VST, BPF are refereed to Mean-shift tracking~\cite{meanshift}, Context tracking~\cite{context}, Visual Silhouette Tracker \cite{CRFVisualSilho} and Boosted Particle Filtering~\cite{particlefilter} respectively. }
     \label{tab:res1}
    \begin{tabular}{c||c|c|c|c|c}
%    \hline
	\textbf{Video Name} & \textbf{MST} ~\cite{meanshift} &\textbf{CT} ~\cite{context} & \textbf{VST} \cite{CRFVisualSilho} & \textbf{BPF}~\cite{particlefilter} & \textbf{DS-CRF} \\ \hline
    \textbf{Subway1} & 88\% & 92\%& 100\% & 100\%&100\%\\
    \textbf{Subway2} & 16\%& 42\%& NA &100\%&100\% \\
     \end{tabular}
     \end{center}

\end{table*}

\begin{figure*}[!h]
\centering{

 \includegraphics[scale = 0.8]{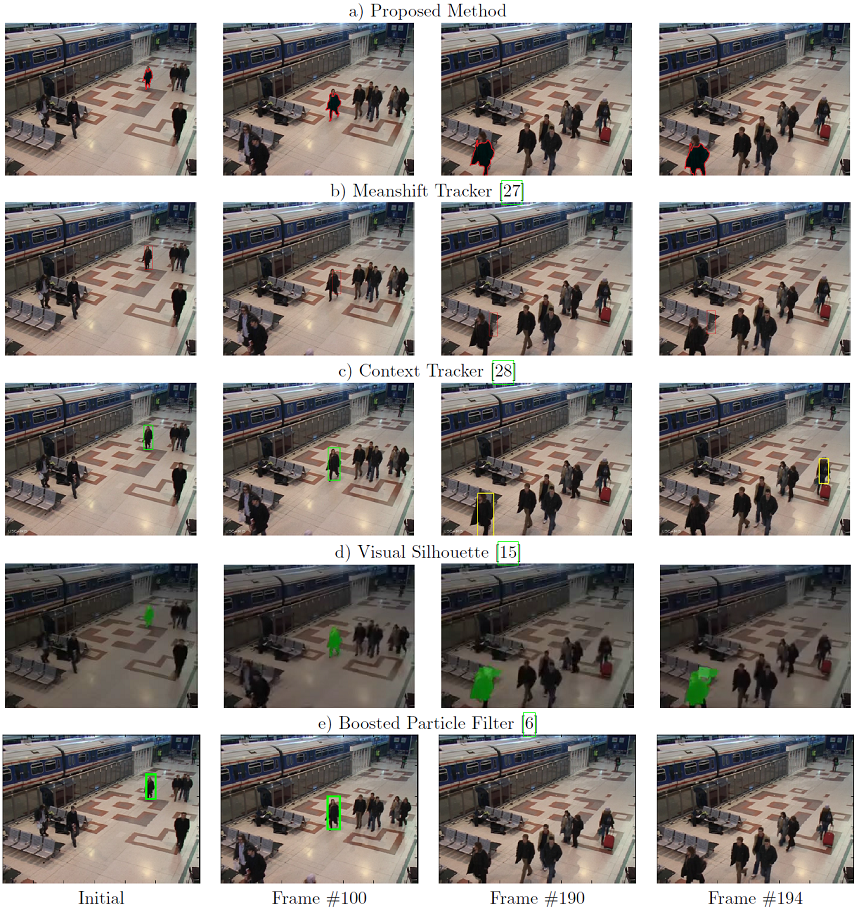}
}
\caption{\textbf{Example tracking results for Subway1.} It can be observed that while the mean-shift tracking, context tracking, and boosted particle filtering methods lose the object target completely, both the visual silhouette tracking method and the proposed DS-CRF method is able to track the object silhouette all the way through.  It can also be observed that the object silhouette obtained using the proposed method is more accurate than that obtained using the visual silhouette method.}
\label{fig:result1}
\end{figure*}

First, let us examine the performance of the proposed DS-CRF method in the situation where the object being tracked changes significantly in size and shape over time.  Fig. \ref{fig:result1} shows the single-object object silhouette tracking results of the tested tracking methods for the Subway1 sequence.   It can be observed that while the mean-shift tracking, context tracking, and boosted particle filtering methods lose the object target completely, both the visual silhouette tracking method and the proposed DS-CRF method is able to track the object silhouette all the way through.  It can also be observed that the object silhouette obtained using the proposed method is more accurate than that obtained using the visual silhouette method.  These results illustrate the capability of the proposed DS-CRF method in tracking the object silhouette over time in spite of drastic changes in size and shape over time.

\begin{figure*}[!h]
\centering{

  \includegraphics[scale = 1.2]{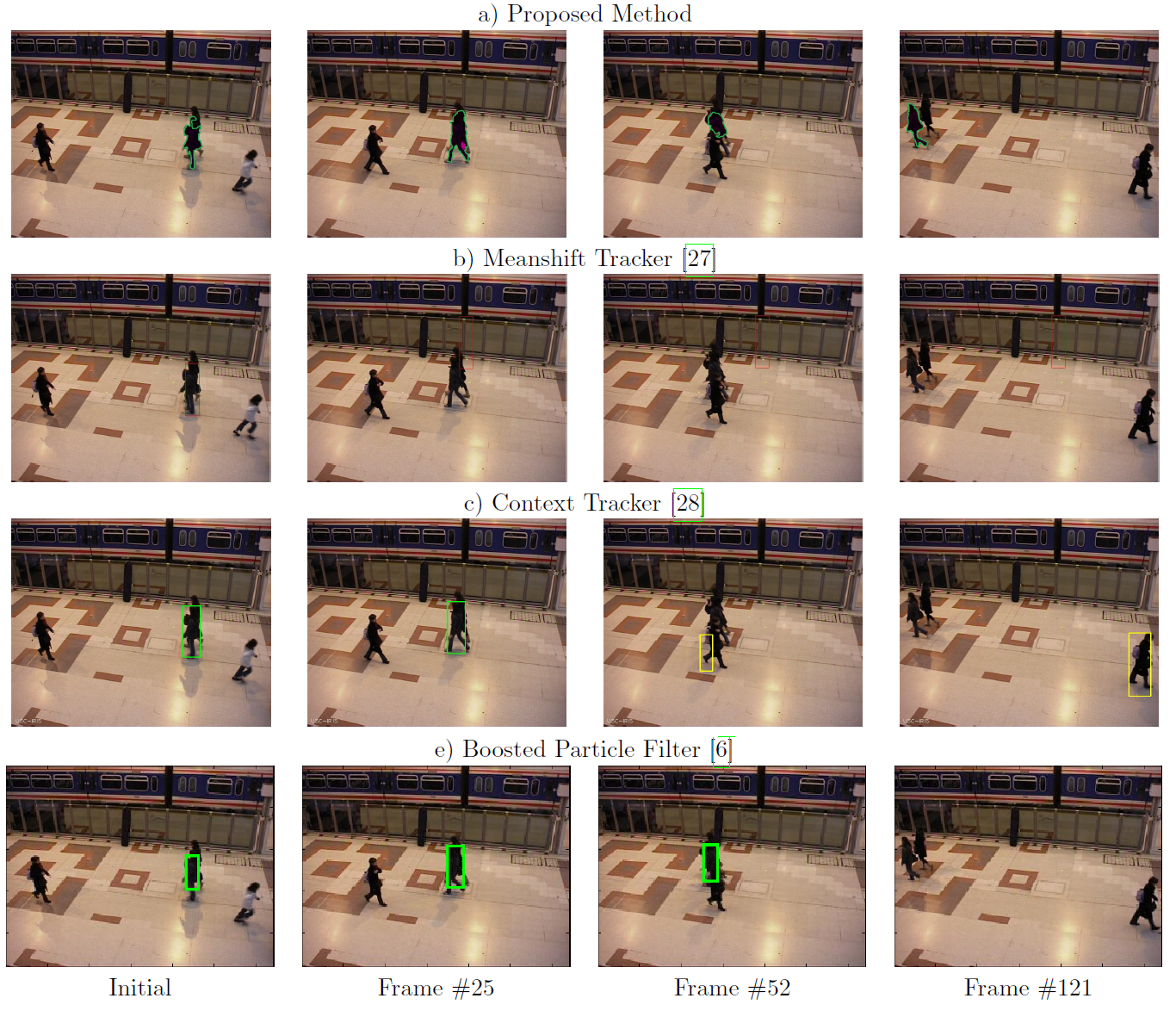}
}
\caption{\textbf{Example tracking results for Subway2.} It can be observed that while the mean-shift tracking, context tracking, and boosted particle filtering methods lose the object target completely, the proposed DS-CRF method is able to track the object silhouette all the way through despite being occluded by another person.}
\label{fig:result3}
\end{figure*}

\begin{table*}[!Htp]
	
	\begin{center}
    \caption{Comparison results for the Subways3 sequence. The accuracy is reported as the average of the accuracy for all tracked targets since this dataset contains three targets. }
     \label{tab:res2}
    \begin{tabular}{c||c|c}
%    \hline
	\textbf{Video Name} & \textbf{Boosted Particle Filtering}~\cite{particlefilter} & \textbf{DS-CRF} \\ \hline
    \textbf{Subway3} & 80\%& 100\%

     \end{tabular}
     \end{center}

%    \vspace{-0.6 cm}
\end{table*}
Next, let us examine the performance of the proposed DS-CRF method in the situation where the object being tracked undergoes occlusion by other objects over time.  Fig. \ref{fig:result3} shows the single-object object silhouette tracking results of the tested tracking methods for the Subway2 sequence.  It can be observed that while the mean-shift tracking, context tracking, and boosted particle filtering methods lose the object target completely, the proposed DS-CRF method is able to track the object silhouette all the way through despite being occluded by another person.  These results illustrate the capability of the proposed DS-CRF method in tracking the object silhouette over time in spite of object occlusion.

\begin{figure*}[!h]

  \includegraphics[scale = 1.2]{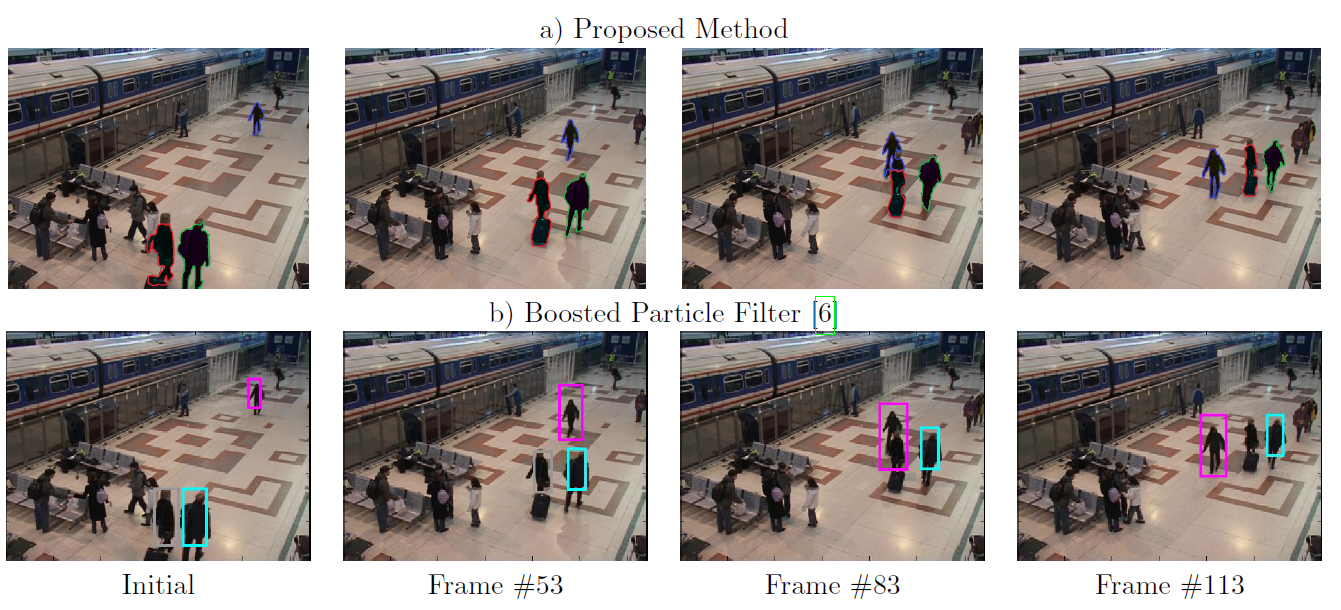}
\caption{\textbf{Example tracking results for Subway3.} It can be observed that while the boosted particle filtering method is able to track two of the three object targets completely, it loses one of the object targets as a result it crossing paths with one of the other object targets.  Furthermore, the boosted particle filtering method does not provide pixel-level object silhouettes and is able to only track bounding boxes.  On the other hand, the proposed DS-CRF method is able to track all three of the object silhouettes at the pixel-level all the way through.  }
\label{fig:result4}
\end{figure*}

Finally, let us examine the performance of the proposed DS-CRF method in the situation where we wish to track multiple object silhouettes over time. Fig. \ref{fig:result4} shows the multiple-object silhouette tracking results of the tested tracking methods for the Subway3 sequence.   It can be observed that while the boosted particle filtering method is able to track two of the three object targets completely, it loses one of the object targets as a result it crossing paths with one of the other object targets.  Furthermore, the boosted particle filtering method does not provide pixel-level object silhouettes and is able to only track bounding boxes.  On the other hand, the proposed DS-CRF method is able to track all three of the object silhouettes at the pixel-level all the way through.  These results illustrate the capability of the proposed DS-CRF method in tracking multiple object silhouettes over time in a reliable manner.

\section*{Discussion}

Here, we proposed a deep-structured conditional random field (DS-CRF) model for object silhouette tracking.  In this model, a series of state layers are used to characterize the object silhouette at all points in time within a video sequence.  Connectivity between state layers formed dynamically based on inter-frame optical flow allows for interactions between adjacent state layers to facilitate for the utilization of both spatial and temporal context within a deep-structured probabilistic graphical model.  Experimental results showed that the proposed DS-CRF model can be used to facilitate for accurate and efficient pixel-level tracking of object silhouettes that can change greatly over time, as well as under different situations such as occlusion and multiple targets within the scene. Experiment results using both simulated data and real-world video datasets containing different scenarios demonstrated the capability of the proposed DS-CRF approach to provided strong object silhouette tracking performance when compared to existing tracking methods.

One of the main contributing factors to the proposed method's ability to handle uncertainties in object motion dynamics and size and shape changes over time is in the way the inter-layer connectivity is established dynamically based on inter-frame optical flow information.  If the inter-layer connectivity is established statically at all state layers of the deep-structured model, then the feature functions would hold little meaningful relationships in the temporal domain as the object accelerates and changes size over time.  By making use of inter-frame optical flow information to determine inter-layer connectivity between adjacent state layers, the feature functions maintain meaning over time in guiding the prediction process.  Another important contributing factor is the incorporation of object shape feature functions (spatial coherency) enforces the proposed method to consider object shape variations in time, which also aids in the handling of changes in size and shape over time.

Future work involves extending the proposed DS-CRF model to incorporate not only inter-frame optical flow information, but also additional motion information via descriptor matching to better guide the establishment of inter-layer connectivity in the situation of large object displacements within a short time in the video sequence.  Furthermore, we aim to explore the extension of the DS-CRF model with high-order and fully-connected clique structures~\cite{SFCRF} to improve modeling of spatial relationships for better object silhouette boundaries.   Finally, we aim to explore the application of the proposed DS-CRF model for the purpose of improved video saliency detection using texture distinctiveness-based feature functions~\cite{saliency1,saliency2,saliency3} and improved content-based video retargeting using energy gradient feature functions~\cite{seamcarving}.

\section*{Supporting Information}

\section*{Acknowledgments}

This work was supported by the Natural Sciences and Engineering Research Council of Canada, Canada Research Chairs Program, and the Ontario Ministry of Economic Development and Innovation.
The financial support provided by Shiraz University for this work is also appreciated.

%\section*{Author contributions}
%
%M.S., Z.A., and A.W. conceived and designed the DS-CRF model.  M.S., Z.A., and A.W. worked on formulation and derivation of solution for the DS-CRF model.  M.S. performed the data processing.  All authors contributed to writing the paper and to the editing of the paper.
%

\pagebreak

%\nolinenumbers

%

%\section*{Tables}
%\begin{table*}[!h]
%%\vspace{-0.3cm}
%\begin{center}
%\caption{\label{ltf} lost target frame. }
%\large {
%\begin{tabular}{|l|c||c|c|c|c|}\hline
%Video Sequence &Frames& Meanshift&Context Tracker & Particle Filter & Ours \\ \hline \hline
%Sequence1 &  208 & 185 & 193 & \color{green}\footnotesize Not Lose &\color{green}\footnotesize Not Lose\\ \hline
%Sequence2 &122 & 20& 52 & \color{green}\footnotesize Not Lose & \color{green} \footnotesize Not Lose \\ \hline
%Sequence3 & 194 &\color{blue} NA & \color{blue} NA &\color{red} \footnotesize Loses One Target@80 & \color{green}\footnotesize Tracks All Targets \\ \hline
%\end{tabular}}
%%\vspace{-0.2cm}
%%\vspace{-0.5cm}
%\end{center}
%\end{table*}

%\section*{Supporting Information Legends}
%
% Please enter your Supporting Information captions below in the following format:
%\item{\bf Figure SX. Enter mandatory title here.} Enter optional descriptive information here.
%
%\begin{description}
%\item {\bf}
%\item {\bf}
%\end{description}

\end{document}